# Ultra-short-term multi-step wind speed prediction for wind farms based on adaptive noise reduction technology and temporal convolutional network

Haojian Huang [1,*]
haojianhuang927@gmail.com

**Abstract**: As an important clean and renewable kind of energy, wind power plays an important role in coping with energy crisis and environmental pollution. However, the volatility and intermittency of wind speed restrict the development of wind power. To improve the utilization of wind power, this study proposes a new wind speed prediction model based on data noise reduction technology, temporal convolutional network (TCN), and gated recurrent unit (GRU). Firstly, an adaptive data noise reduction algorithm P-SSA is proposed based on singular spectrum analysis (SSA) and Pearson correlation coefficient. The original wind speed is decomposed into multiple subsequences by SSA and then reconstructed. When the Pearson correlation coefficient between the reconstructed sequence and the original sequence is greater than 0.99, other noise subsequences are deleted to complete the data denoising. Then, the receptive field of the samples is expanded through the causal convolution and dilated convolution of TCN, and the characteristics of wind speed change are extracted. Then, the time feature information of the sequence is extracted by GRU, and then the wind speed is predicted to form the wind speed sequence prediction model of P-SSA-TCN-GRU. The proposed model was validated on three wind farms in Shandong Province. The experimental results show that the prediction performance of the proposed model is better than that of the traditional model and other models based on TCN, and the wind speed prediction of wind farms with high precision and strong stability is realized. The wind speed predictions of this model have the potential to become the data that support the operation and management of wind farms. The code is available at link.

## 1 Introduction

With the vigorous development of green energy development technology, clean energy represented by wind power has become more and more concerned. With the benefits of low construction cost, few restrictions, and no pollution, wind power has become the fastest growing renewable energy source in the world (Lin Z et al. 2020; He X et al. 2023). However, due to the intermittent, unstable, and nonlinear characteristics of wind, the efficient utilization of wind power has become a major issue. On the one hand, the large fluctuation of wind speed does not take advantage of the safe and stable operation of large-scale power grids. On the other hand, the distribution of wind speed in different periods is not balanced, which also makes it difficult to maintain a high level of utilization and a low level of cost through a reasonable dispatch (Sun W et al. 2018; Zhang D et al. 2019). But if the ultra-short-term prediction of wind speed with excellent enough accuracy is realized, these problems will be significantly alleviated. Therefore, ultra-short-term wind speed prediction has become a profound issue in the field of wind power development.

Over the past few years, a number of wind speed prediction models have been proposed to realize prediction. In terms of the working principle of them, four types of models are mainly included, which are physical (Yang M et al. 2023), statistical (Szarek D et al. 2023), spatiotemporal correlation (Baïle R, Muzy J F 2023) and artificial intelligence models (Zhang S et al. 2020; Ping Jiang et al. 2021; Yousef Ali, Hamed H. Aly 2024). The physical model predicts wind speed by considering physical processes in the atmosphere in terms of numerical weather prediction (NWP). Factors including air temperature, humidity, and altitude will be taken into account, which leads to high complexity and time cost in its calculation process (Yang M et al. 2023; Sankar S R, Panchapakesan M 2024). As a result, it can not be applied for short-term wind speed prediction . In contrast, statistical models, such as the autoregressive moving average (ARMA) model (Shi J et al. 2011), the autoregressive moving average (ARIMA) model(Singh S et al. 2020), and the fractional-ARIMA model(Kavasseri RG et al. 2009), are applied in short-term wind speed predictions usingvarieties of statistical models.. However, it is difficult to keep them woring effectively for a long time as they are inseparable from priori assumptions. The spatiotemporal correlation model is based on the temporal and spatial correlations between the target and neighboring sites. However, its overwhelming complexity and implementation costs limit the performance in the short-term wind speed prediction (Yong Chen et al. 2019).

With the gradual prosperity of artificial intelligence, a large number of prediction models based on machine learning are gradually applied to achievewind speed prediction with high accuracy. Specifically, these models can be divided into shallow learning models and deep learning models (Niu D et al. 2022). Shallow learning models include artificial neural networks (Karasu S et al. 2017; Nielson J et al. 2020; Yousef Ali, Hamed H. Aly 2024), extreme learning machines(Li N et al. 2019; Wang J et al. 2024), and support vector machines(Karasu S et al. 2017). They were once widely concerned by scholars for their superior nonlinear feature learning ability. However, shallow learning models are prone to falling into local optima, overfitting, or poor convergence, which limits their prediction accuracy(Khodayar M et al. 2018). In contrast, deep learning models fundamentally adapt to the characteristics of sequence prediction with their superior network structures(Wang H et al. 2019; Nascimento E G S 2023). Among them, the recurrent neural network (RNN) makes prediction more effective by preserving the temporal relationship between variables (Chung J et al. 2014; Zn A et al. 2021; Zheng J, Wang J 2024). To further solve the problem of RNN gradient vanishing, the long short-term memory (LSTM) network was proposed and achieved high prediction accuracy (Shahid F et al. 2021; Han L et al. 2019; Liu X, Zhou J 2024). Considering that very few wind speed sequence samples are collected in ultra-short-term wind speed prediction (Guan S et al. 2024), the gated recurrent network (GRU) greatly improves the convergence speed of training by simplifying the network structure of LSTM while achieving high-accuracy prediction (Wang Y et al. 2018; Venkatachalam K et al. 2023; Liu Z H, et al. 2024; Yu H, et al. 2024). As the convolutional neural network (CNN) is famous for its superior feature extraction ability in image processing, its potential to be applied in wind speed prediction has received extensive attention from scholars. For example, Zhang J et al. (2021) combined CNN and LSTM to extract key features of wind farm clusters, and provided a new high-precision method for wind farm cluster power prediction. Although the methods based on CNN above have made a difference for improving the prediction accuracy (Zhang Y M, Wang H 2023; Joseph L et al. 2024), the time and computer memories for models' training is quite overwhelming. Bai S et al. (2018) proposed a new network called temporal convolutional network (TCN), which has the advantages of both CNN and RNN, and experiments show that TCN outperforms the LSTM model in prediction accuracy (Lara-Benítez P et al. 2020; Gong M, et al. 2023; Zhang G et al. 2024). And fortunately, some studies have demonstrated that TCN with unique structures outperforms LSTM networks when dealing with long-step time series, indicating that TCN can effectively utilize convolutional methods to extract the features in complex time series, which makes it suitable for forecasting ultra-short-term wind speed prediction (Zha W et al. 2022). In addition to using TCN to extract features from complex wind speed sequences, the sequence decomposition method has also been proved to be an effective means. They decompose and reconstruct complex wind speed sequences and extract useful features from subsequences. By combining the AI-based models into a hybrid model, the prediction accuracy can be significantly improved (Jiang Z, et al. 2023; Zhao Z et al. 2023; Gong Z et al. 2024; Sankar S R, Panchapakesan M 2024). Generally, empirical mode decomposition (EMD) (Liang Y et al. 2019), ensemble empirical mode decomposition (EEMD) (Santhosh M et al. 2018) and variational mode decomposition (VMD) (Zhang G et al. 2019; Han L et al. 2019; Zhao Z et al. 2023) are common sequence decomposition methods. However, the methods based on the empirical mode decomposition framework often suffer from the risk of introducing virtual modes (Xiong D et al. 2021; Xiong D et al. 2019; Xiao Y et al. 2023; Huang H 2024). Although variational mode decomposition does not introduce virtual modes like the former, one of its important hyperparameters $K$ (the number of intrinsic mode functions, IMF) needs to be determined in advance (Duan J et al. 2021). And this problem also arises in decomposition methods based on singular spectrum analysis (SSA).

*Corresponding author

To solve the above problems, this paper uses the Pearson coefficient to search for the best hyperparameter for singular spectrum analysis to reinforce the SSA algorithm, and it is called P-SSA. Meanwhile, the TCN is combined with GRU so that a novel robust P-SSA-TCN-GRU hybrid model for ultra-short-term multi-step wind speed prediction is proposed. Specifically speaking, P-SSA is applied to denoise the original sequence in the data preprocessing step. Next, feature extraction is performed in combination with the TCN model. Finally, based on the feature sequence, GRU is used to predict the wind speed sequence. The main contributions of this paper are summarized below:

(i). Considering the effectiveness of singular spectrum analysis in sequence noise reduction, it is proposed to further optimize the denoising effect of SSA by Pearson coefficient.

(ii). Taking the effectiveness of the hybrid model in predicting complex wind speed sequences into account, GRU will be combined with TCN to achieve feature extraction and high-precision prediction of complex wind speed sequences.

(iii). A novel robust hybrid prediction model (P-SSA-TCN-GRU) is proposed for ultra-short-term multi-step wind speed forecasting. And experiments and comparison analyses are performed to fully evaluate the validity of the proposed model.

## 2 Methodology

The methods and theories applied in this paper will be briefly introduced in this section before proposing the P-SSA-TCN-GRU hybrid prediction model.

### 2.1 Singular spectrum analysis

SSA is used to decompose original sequences into interpretable subsequences. The process can be summarized as follows:

**Step1** (Embedding): The original time series $c(c_1, c_2, \cdots, c_N)$ is converted into sequence $t(t_1, t_2, \cdots, t_K)$ through Equation (1).

$$c(c_1, c_2, \cdots, c_N) \to t(t_1, t_2, \cdots, t_K) \tag{1}$$

Then, the dimension embedded in SSA is denoted by $S$. For $K = N - S + 1$, $S$ lag vector can be defined as: $t_i(c_i, c_{i+1}, \cdots, c_{i+S-1})^T \in R^S$, $S \in [2, N]$. And $t = (t_1, t_2, \cdots, t_K)$ is expressed as the trajectory matrix in Equation (2).

$$t = (t_1, t_2, \cdots, t_K) = \begin{pmatrix} c_1 & c_2 & \cdots & c_K \\ c_2 & c_3 & \cdots & c_{K+1} \\ \cdots & \cdots & \cdots & \cdots \\ c_S & c_{S+1} & \cdots & c_N \end{pmatrix} \tag{2}$$

**Step2** (Singular value decomposition, SVD): For the covariance matrix $s = t \cdot t^T$, the singular value decomposition method is used to generate $S$ eigenvalues $(\lambda_1, \lambda_2, \cdots, \lambda_S)$ and eigenvectors $(u_1, u_2, \cdots, u_S)$. If $z = \max(i, such\ that\ \lambda_i > 0)$ and $v_i = t^T u_i \sqrt{\lambda_i}$ $(i = 1,2, \cdots, h)$, SVD of the trajectory matrix can be expressed as:

$$\boldsymbol{t} = \boldsymbol{t_1} + \boldsymbol{t_2} + \cdots + \boldsymbol{t_h} \tag{3}$$

where $t_i = \sqrt{\lambda_i} u_i v_i$, with a rank of 1, and its main component $v_1, v_2, \cdots, v_h$. Besides, $t_i$ is the eigen solution of SVD of $t$.

**Step3** (Grouping): The range is decomposed into subsets $a_1, a_2, \cdots, a_m$; and there is no connection among them. The resulting matrix $y_a$ of $a = (n_1, n_2, \cdots, n_p)$ can be expressed as $y_a = y_{a_1} + y_{a_2} + \cdots + y_{a_p}$, and the trajectory matrix is decomposed into $y = y_{a_1} + y_{a_2} + \cdots + y_{a_m}$.

**Step4** (Diagonal averaging): The primary purpose of this process is to convert each resulting matrix $y_a$ into a new series of length $N$. Assuming $d$ is a matrix of $S \times K$, where $S^* = \min(S, K)$ and $K^* = \max(S, K)$. If $S < K$, then $y^*_{ij} = y_{ij}$; otherwise $y^*_{ij} = y_{ji}$. Then transform $y$ to a sequence $(r_1, r_2, \cdots, r_S)$ according to the following equation:

$$r_k = \begin{cases} \dfrac{1}{k+1} \sum_{q=1}^{k+1} y^*_{q,k-q+1}, 1 \le k \le S^* \\ \dfrac{1}{S^*} \sum_{q=1}^{S^*} y^*_{q,k-q+1}, S^* \le k \le K^* \\ \dfrac{1}{N-K+1} \sum_{q=k-K^*+1}^{N-K^*+1} y^*_{q,k-q+1}, K^* \le k \le N \end{cases} \tag{4}$$

Therefore, the reconstructed sequence can be obtained. In practical applications, if the first $m$ $(m < L)$ components with larger eigenvalues are selected, the effect of filtering noise can be achieved.

### 2.2 Temporal convolutional network

Convolutional networks have been shown to perform well in extracting high-level features from structured data. The temporal convolutional network (TCN) is a neural network model that utilizes causal convolution and dilated convolution, which can adapt to the time-series nature of time-series data and provide a field of view for time-series modeling. It forms residual blocks by stacking multiple layers of dilated convolutions and connecting them through residuals; then multiple residual blocks are used to extract features, and the residual blocks are connected by Skips; activated by the ReLU activation function, and output by Softmax. The schematic diagram of the TCN structure is shown in the following figure.

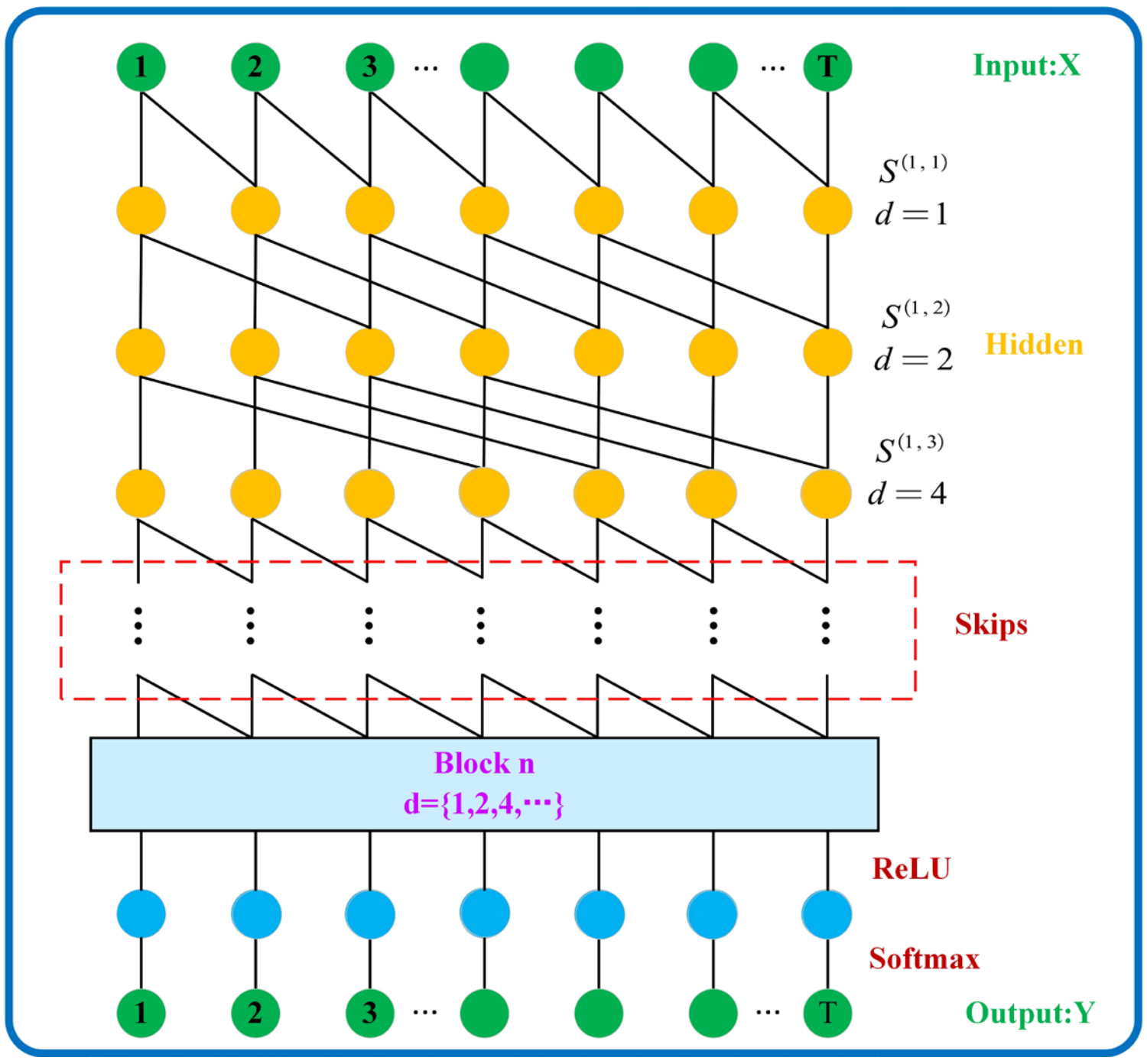

**Fig. 1** The structure of TCN

Defined $S_t^{(j,l)}$ as the state of the $j^{\text{th}}$ dilated convolution in the $l^{\text{th}}$ residual block of the network at $t$. The input of the residual block $S^{(j,l)}$ is the same as $S^{(j-1,L)}$ , the input $S^{(1,1)}$ is a single sample when $j = 1$ and $L$ is the number of dilated convolutional layers in each residual block. Then $\widehat{S}_t^{(j,l)}$ is the output of the dilated convolution at time , and $S_t^{(j,l)}$ is the output after adding residual connections, which is calculated as follows:

$$\boldsymbol{S_t^{(j,l)} = f(W_1 \cdot S_{t-d}^{(j,l-1)} + W_2 \cdot S_t^{(j,l-1)} + b)} \tag{5}$$

$$\boldsymbol{S_t^{(j,l)} = S_t^{(j,l-1)} + V \cdot \widehat{S}_t^{(j,l)} + e} \tag{6}$$

Where $W = \{W_1, W_2\}$, $V$ are the weights of the dilated convolution; $b$ and $e$ are the biases of the dilated convolution. And in each layer, $\{W, V, b, e\}$ of the dilated convolution is different.

After completing each layer of dilated convolution operations, compute the output $Z_t^{(0)}$ of the skip-connected residual block:

$$\boldsymbol{Z_t^{(0)} = ReLU(\sum_{j=1}^{B} S_t^{(j,L)})} \tag{7}$$

Where $B$ is the number of the residual block. Then calculate the hidden layer state $Z_t^{(1)}$:

$$\boldsymbol{Z_t^{(1)} = ReLU(U_r \cdot Z_t^{(0)} + r)} \tag{8}$$

Where $U_r$ and $r$ is the hidden layer weight and bias, respectively. Finally, the prediction is output through Softmax activation:

$$\widehat{Y_t} = \boldsymbol{Softmax}(\boldsymbol{U} \cdot \boldsymbol{Z}_t^{(1)} + \boldsymbol{c}) \tag{9}$$

Where $U$ and c is the hidden layer weights and bias respectively.

**2.3 Gated recurrent unit network**

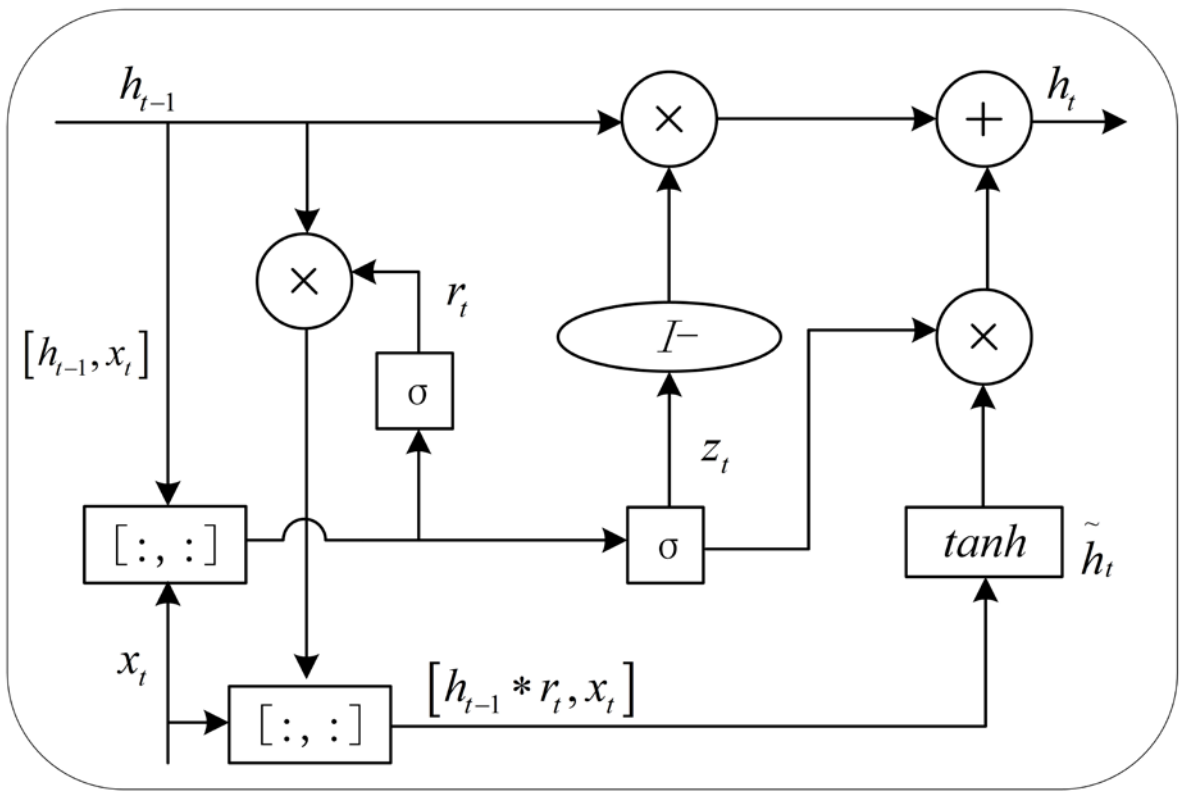

**Fig. 2** The structure of GRU

GRU network is an improved model of long short-term memory network. It optimizes the three gate functions of the long short-term memory network, integrates the forget gate and the input gate into a single update gate, and mixes the neuron state and the hidden state. This can effectively alleviate the "gradient vanishing" problem in the recurrent neural network, reduce the number of parameters of the long short-term memory network unit, and shorten the training time of the model. The basic structure of the network is shown in **Fig. 2**, and the mathematical description is shown in Equation (10).

$$\begin{cases} r_t = \sigma(W_r \cdot [h_{t-1}, x_t]) \\ z_t = \sigma(W_z \cdot [h_{t-1}, x_t]) \\ \tilde{h} = tanh(W_{\tilde{h}} \cdot [r_t \times h_{t-1}, x_t]) \\ h_t = (I - z_t) \times h_{t-1} + z_t \times \tilde{h}_t \\ y_t = \sigma(W_o \cdot h_t) \end{cases} \tag{10}$$

In **Fig. 2** and Equation (10), $x_t$, $h_{t-1}$, $h_t$, $r_t$, $z_t$, $\tilde{h}_t$, $y_t$ are the input vector, the state memory variable at the previous moment, the state memory variable moment at the current moment, the update gate's state, the state of the reset gate, the state of the current candidate set, and the output vector at the current moment. $W_r$, $W_z$, $W_{\tilde{h}}$, $W_o$ are the weight parameters of the connection matrix formed by multiplying $x_t$ and $h_{t-1}$ by the update gate, reset gate, candidate set, and output vector respectively; $I$ represents the identity matrix; $\sigma$ represents the activation function; tanh is the tangent function.

The core module of the GRU network is the update gate and reset gate. The split matrix of the input variable $x_t$ and the state record variable $h_{t-1}$ at the previous moment is input into the update gate through a nonlinear transformation with sigmoid, which determines the degree to which the state variable at the previous moment is brought to the current state. The reset gate controls the amount of information that was last written into the candidate set. The information of the last time is stored in $I - z_t$ times of $h_{t-1}$, and the information of the current time is recorded in $\tilde{h}_t$ times of $z_t$. Finally, their sum is the output of the current time.

**2.4 Integrated model framework**

The model commences with the deployment of SSA to the raw wind sequence. This pivotal preprocessing step, guided by the Pearson correlation coefficient, filters out noise and extraneous components, yielding a refined sequence. Through embedding and reconstruction, the model transforms the initial series $c(c_1, c_2, \cdots, c_N)$ into a denoised series $t(t_1, t_2, \cdots, t_K)$, maintaining the fundamental dynamics while excluding volatilities that obfuscate the prediction process.

Following denoising, the TCN coupled with GRU is tasked with learning from the historical context encapsulated within the wind speed series. The TCN's architecture, through dilated convolutions, extracts temporal features and identifies patterns over multiple time scales. The convolutional layers capture both local and long-range dependencies through Equation (5).

Subsequently, the GRU layers leverage their gating mechanisms to manage the flow of information, effectively allowing the network to learn from long sequences while mitigating the risk of gradient vanishing. The GRU's update and reset gates to adapt the memoey content, balancing the retention of prior knowledge with the acquisition of new latent information.

The model's output layer integrates the learned features into a predictive result through a Softmax activation function, generating a probability distribution over possible future wind speeds. This prediction obtained with Equation (10) serves as the final estimation of imminent wind conditions.

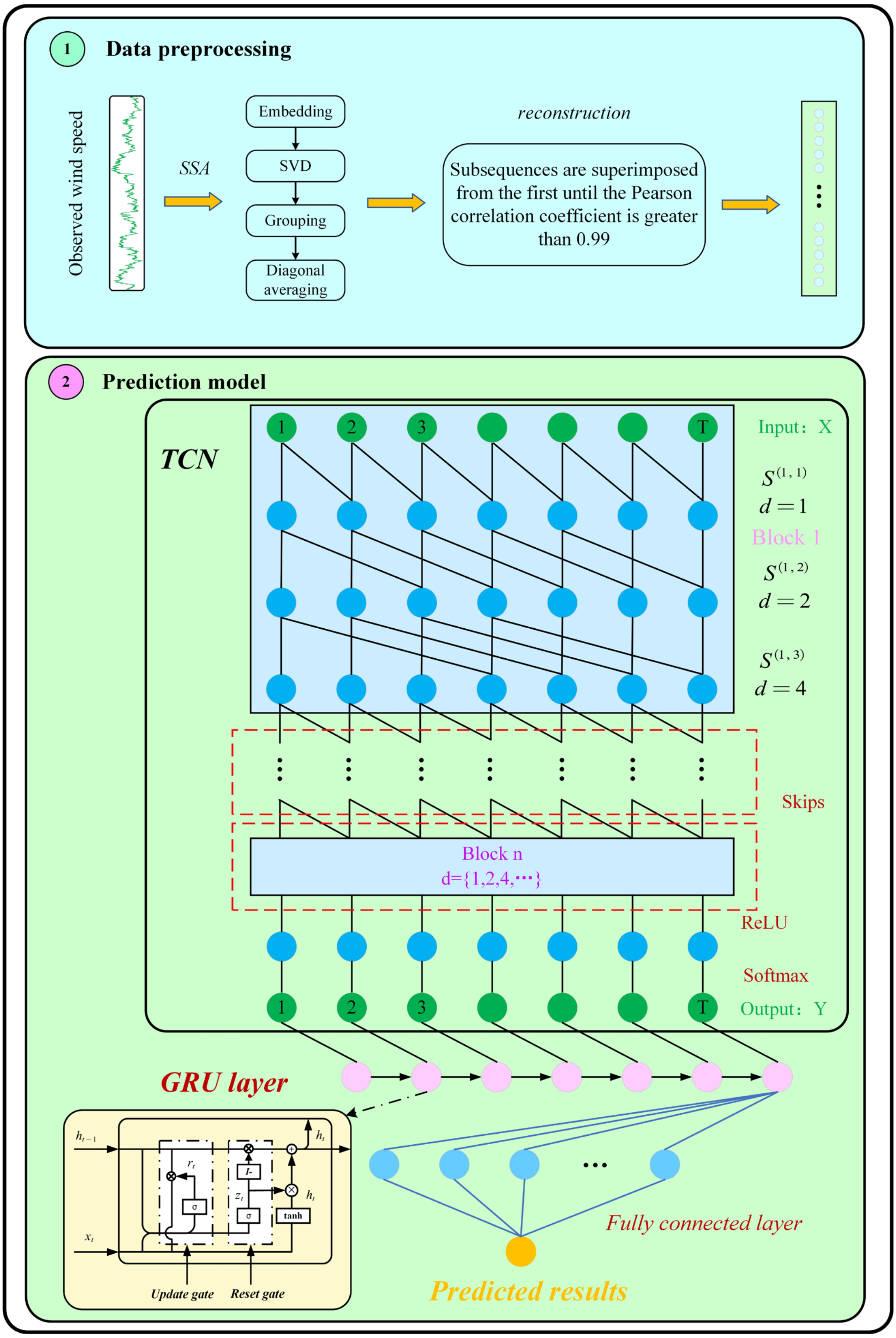

**Fig. 3** The framework of the proposed model

## 3 Experiment and Discussion

### 3.1 Data Description

Shandong Province is a coastal province in East China, located between the north latitude and east longitude on the east coast of China. The central mountainous area of the province is protruding, the southwest and northwest are low-lying and flat, and the east is gently undulating. The terrain is dominated by mountains and hills. Shandong Province has a vast territory with a long coastline of 3000km long, and the terrain protruding to the Yellow Sea Peninsula breeds abundant wind power resources.

As for the dataset, the wind speed data of three wind farms in Penglai District, Zhaoyuan City, Yantai City, Shandong Province was applied in the experiments. And their sampling time of wind speed data is from 2011.1.2 to 2011.1.21, a total of 20 days. The sampling interval is 10 minutes with a total of 2880 samples. The last 200 samples were selected as the test set to verify the prediction performance of the proposed model while the other samples were used as the training set. To obtain longer-term prediction, multi-step wind speed forecast was performed. The statistics of the experimental dataset are shown in **Fig. 4** and **Table 1**.

**Table 1** The statistical indicators of wind speed in the three wind farms

| Site | Sampling Time | Statistical Indicators | | | |
|---|---|---|---|---|---|
| | | **Mean(m/s)** | **Std.(m/s)** | **Min.(m/s)** | **Max.(m/s)** |
| $S_1$ | 2011.1.2-2011.1.21 | 9.0980 | 2.8463 | 2.3000 | 17.5000 |
| $S_2$ | | 9.2334 | 3.1761 | 2.0000 | 18.0000 |
| $S_3$ | | 9.1001 | 2.8680 | 1.9000 | 17.2000 |

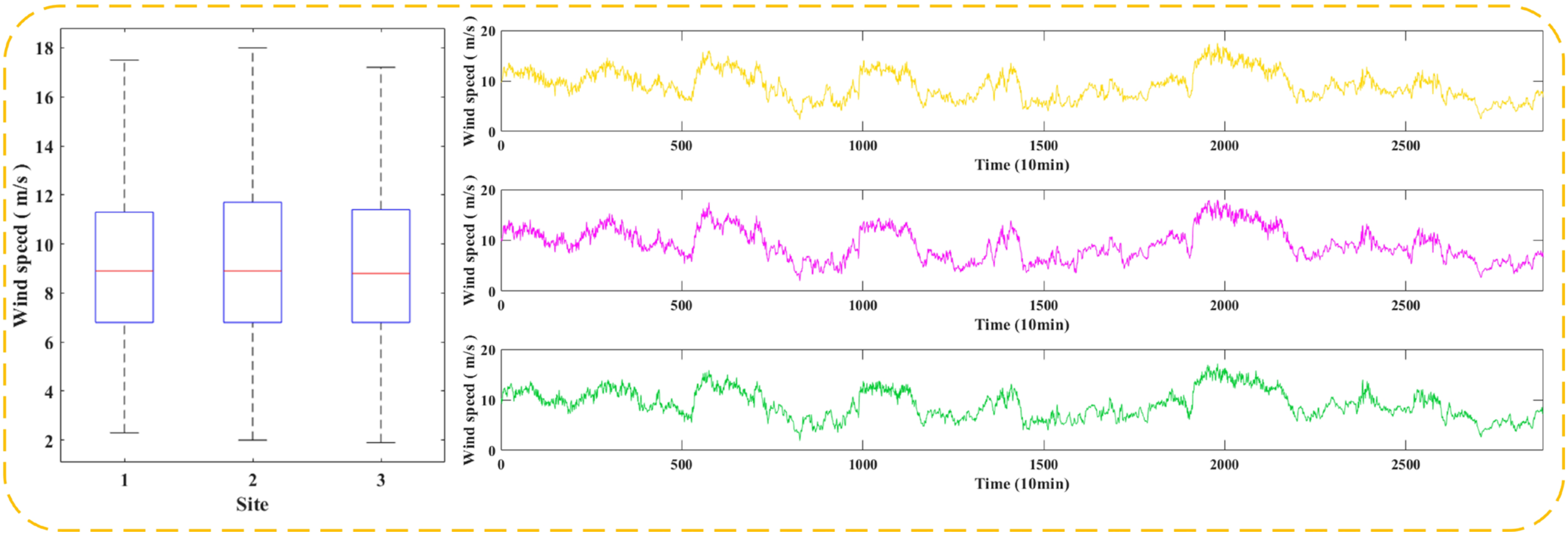

**Fig. 4** Wind speed data in the wind farms

### 3.2 Parameters Setting

When conducting comparative experiments, different parameters settings will greatly affect the prediction accuracy of the model. In terms of the actual situation of the model, mountains of references to other research literature were performed for trial and error. As a result, the parameter settings of the models involved in this work were obtained and shown in **Table 2**.

**Table 2** Parameters setting in the experiment

| Model | Parameters | Value |
|---|---|---|
| TCN | No. of kernel | 10 |
| | Kernel size | 2 |
| | Dilations | [1,2,4] |
| | Optimizer | Adam |
| | Loss | mse |
| | Epochs | 60 |
| | Batch size | 32 |
| GRU/LSTM/RNN | Number of neurons | 64 |
| | Optimizer | Adam |
| | Loss | mse |
| | Epochs | 60 |
| | Batch size | 32 |
| BPNN | Layers | 4 |
| | No. of neurons | [20,20,20,1] |
| | Optimizer | Adam |
| | Loss | mse |
| | Epochs | 60 |
| | Batch size | 32 |
| SSA | Embedded dimension | 15 |
| EMD/EEMD | --- | --- |
| VMD | Alpha | 5000 |
| | No. of decomposition | 15 |
| PSR | Embedded dimension | 20 |
| | Time delay | 1 |

### 3.3 Model Metrics

To quantitatively and comprehensively describe the forecasting accuracy of different forecasting models, three commonly used performance metrics were used to evaluate the forecasting performance of forecasting models, including mean absolute error (MAE), mean absolute percentage error (MAPE), and root mean square error (RMSE). The calculation formula of the evaluation index is shown in **Table 3**, where $N$ is the sequence length $y$ and $\hat{y}$ is the actual value and prediction respectively.

**Table 3** The calculation rules of the metrics

| Metric | Equation | Definition |
|---|---|---|
| MAE | $MAE=\frac{1}{N}\sum_{i=1}^{N}\lvert y_i-\hat{y}_i\rvert$ | The average of absolute error |
| MAPE | $MAPE=\frac{1}{N}\sum_{i=1}^{N}\left\lvert\frac{y_i-\hat{y}_i}{y_i}\right\rvert\times 100\%$ | The average of absolute percentage error |
| RMSE | $RMSE=\sqrt{\frac{1}{N}\sum_{i=1}^{N}(y_i-\hat{y}_i)^2}$ | The root mean square of prediction error |

### 3.4 Experiment I: Comparison of four different noise reduction methods

This experiment aims to verify the performance of the proposed noise reduction method. In the experiment, three noise reduction methods based on the Pearson correlation coefficient, P-EMD, P-EEMD, and P-VMD, are used to compare with the proposed P-SSA, and the prediction model is TCN-GRU. The comparison results between the proposed model and other models are shown in **Fig. 5** and **Table 4**. On the site $S_1$, the MAE, MAPE, and RMSE of the proposed model are 0.0940, 1.9042, and 0.1246, respectively, which are the best among all models. On the site $S_2$, the proposed model has a MAE of 0.0752, which is 0.1161, 0.1630, and 0.1349 lower than the other three models, and this model has the highest prediction accuracy. On the site $S_3$, the RMSE of the proposed model is 0.1220, which is 68.7%, 70.9%, and 50.4% lower than other models, respectively. The performance of this model is still the best.

**Remark:** According to the experimental results in **Table 4**, it can be concluded that the proposed model based on the P-SSA noise reduction algorithm outperforms other models with regard to the accuracy.

**Table 4** Comparison of prediction results based on different noise reduction methods

| **Site** | **Model** | **Denoising Method** | **Metrics** | | |
|---|---|---|---|---|---|
| | | | **MAE** | **MAPE** | **RMSE** |
| $S_1$ | TCN-GRU | P-EMD | 0.2611 | 5.1897 | 0.3254 |
| | | P-EEMD | 0.2877 | 5.5827 | 0.3595 |
| | | P-VMD | 0.1690 | 3.4739 | 0.2171 |
| | | P-SSA | **0.0940** | **1.9042** | **0.1246** |
| $S_2$ | TCN-GRU | P-EMD | 0.1913 | 3.8426 | 0.2138 |
| | | P-EEMD | 0.2382 | 3.6466 | 0.2520 |
| | | P-VMD | 0.2101 | 4.2436 | 0.2633 |
| | | P-SSA | **0.0752** | **1.5659** | **0.1067** |
| $S_3$ | TCN-GRU | P-EMD | 0.3070 | 6.1567 | 0.3907 |
| | | P-EEMD | 0.3350 | 6.4391 | 0.4204 |
| | | P-VMD | 0.1938 | 3.8016 | 0.2462 |
| | | P-SSA | **0.0943** | **1.7905** | **0.1220** |

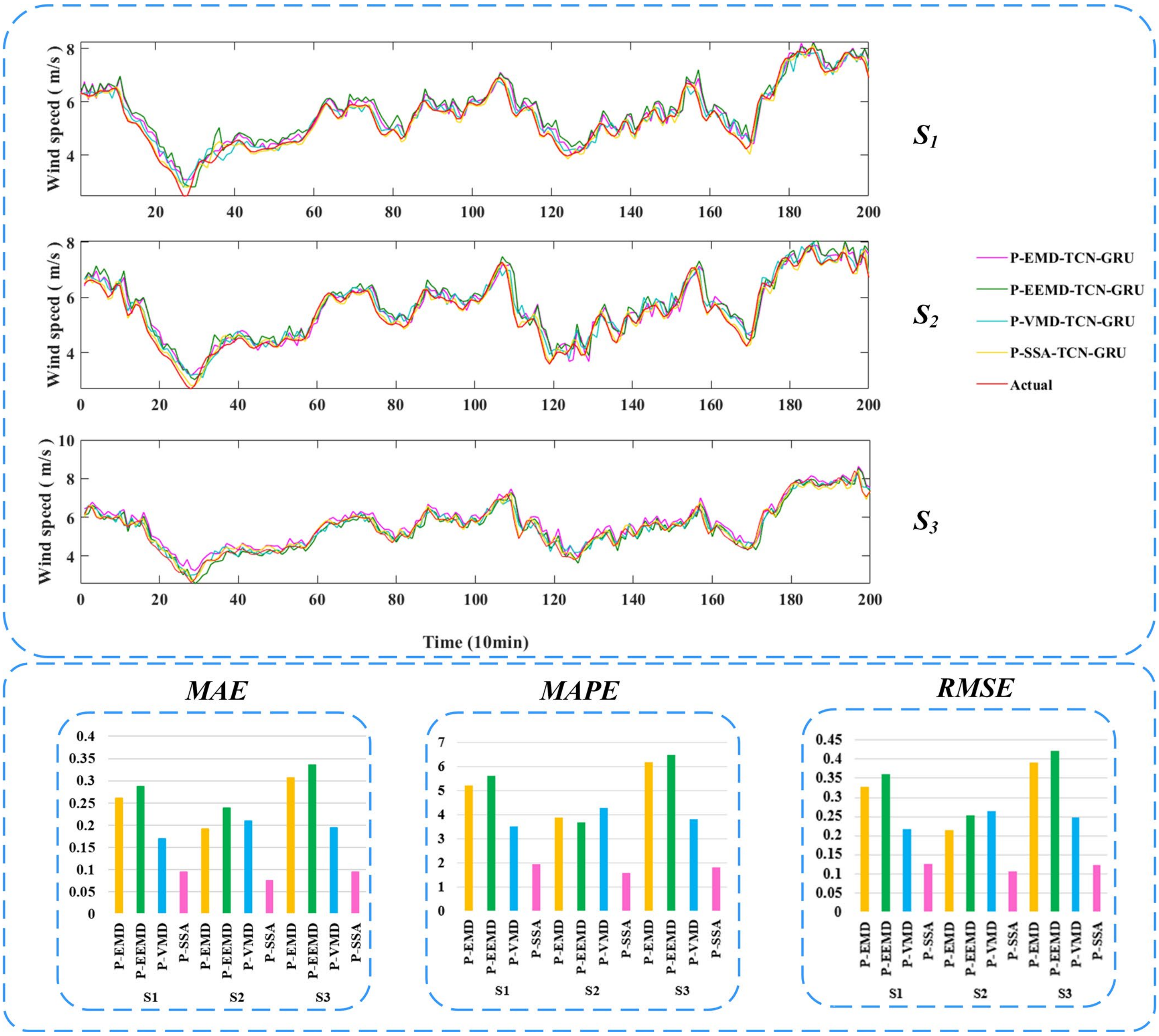

**Fig. 5** The results of four noise reduction methods based on TCN-GRU on three wind farms

### 3.5 Experiment II: Comparison with traditional prediction models

Under the condition that the noise reduction algorithm adopts P-SSA, the proposed TCN-GRU prediction model was compared with GRU, LSTM, RNN, and BPNN to verify the prediction performance of TCN-GRU. The multi-step advance prediction experiments were carried out on the three wind farms $S_1$, $S_2$, and $S_3$. The prediction results of the proposed model and other models based on P-SSA are shown in **Fig. 6** and **Table 5**.

According to the experimental results, in the one-step advance prediction of the $S_1$ site, the prediction error of the proposed model is the smallest: MAE, MAPE, and RMSE are 0.0940, 1.7362%, and 0.1238, respectively. The prediction accuracy of other models from high to low is LSTM, RNN, GRU, and BPNN. In the one-step advance prediction of the $S_2$ site, the prediction accuracy of the proposed model is still the highest: MAE, MAPE, and RMSE are 0.0752, 1.5659%, and 0.2742, respectively, and the prediction error of BPNN is the largest, and the MAPE is 3.4107%. The prediction accuracy of the TCN-GRU model proposed in the one-step advance prediction of the S3 site remains the highest. In the two-step advance prediction, the proposed model has the highest prediction accuracy on $S_1$, $S_2$, and $S_3$ wind farms, with MAPEs of 4.8267%, 5.3426%, and 5.0628%, respectively. The MAPE of the GRU prediction model on the $S_1$, $S_2$, and $S_3$ wind farms is 5.3506%, 5.7411%, and 5.7309%, respectively, which is an increase of 0.5239%, 0.3985%, and 0.6681%, respectively, compared with the TCN-GRU prediction model in the MAPE, which also shows that the performance of the TCN-GRU prediction model is excellent. In the three-step advance prediction, the RMSE of the prediction model of TCN-GRU in $S_1$, $S_2$, and $S_3$ are 0.5294, 0.5791, and 0.5444, respectively, and the prediction performance is better than the traditional prediction models GRU, LSTM, RNN, and BPNN. In the 1-3 step advance prediction of $S_1$ wind farm, the MAPEs of TCN-GRU are 1.7362%, 4.8267%, and 8.0757% respectively. It can be seen that with the increase of the prediction step size, the prediction accuracy of the model decreases, and the prediction error increases.

**Remark**: It can be seen from the 1-3 step advance prediction of the three wind farms $S_1$, $S_2$, and $S_3$ that the prediction performance of the proposed model perform best, which proves that the proposed model is effective in wind speed prediction. Better prediction performance than GRU, LSTM, RNN, and BPNN. It is worth noting that as the prediction step size increases, the model prediction accuracy decreases generally.

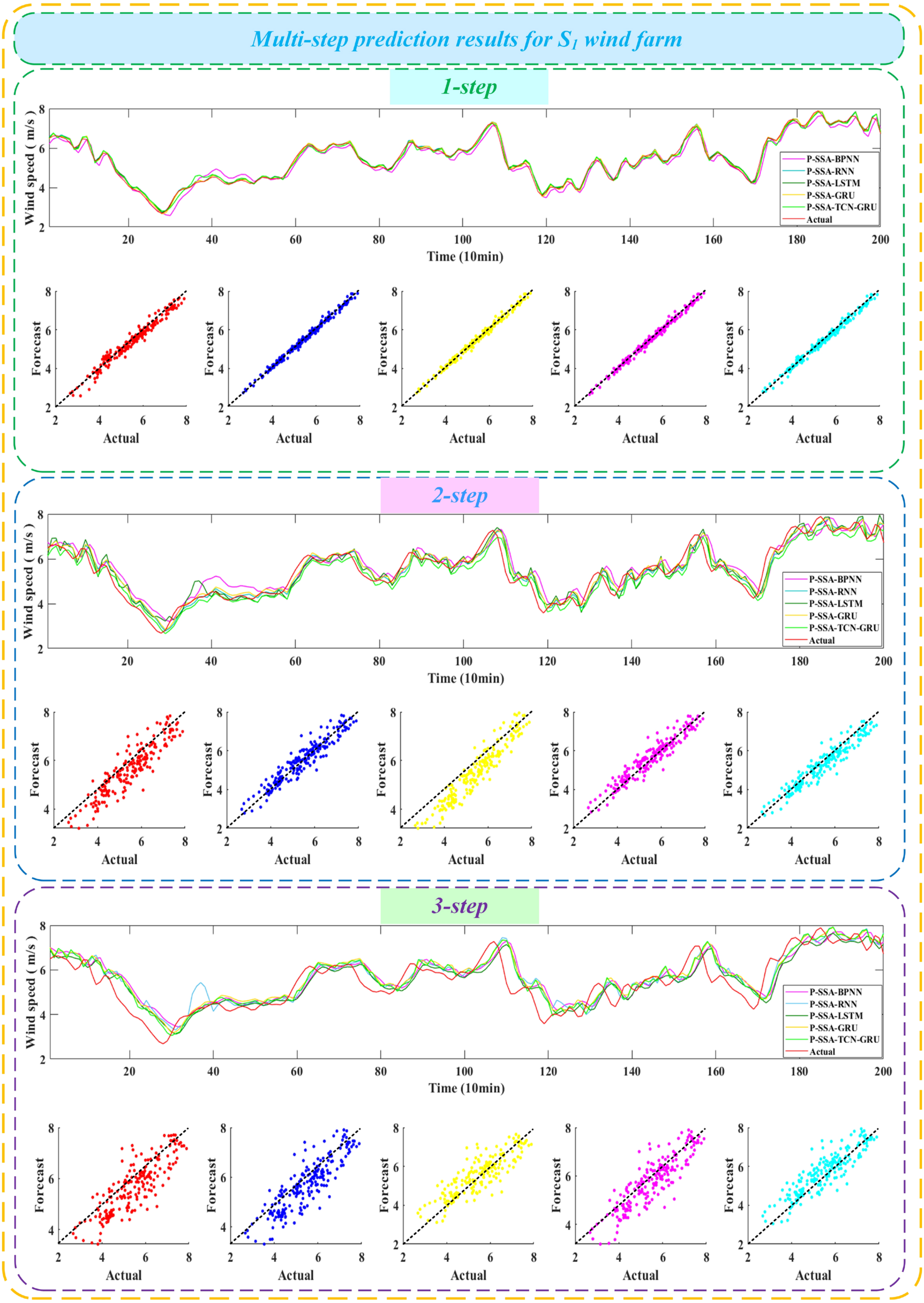

**Fig. 6** The results of the proposed model and the traditional model in the 1-3 steps ahead of the $S_1$ site

**Table 5** Comparison of the prediction results with the traditional models

| Site | Model | 1-step | | | 2-step | | | 3-step | | |
|---|---|---|---|---|---|---|---|---|---|---|
| | | MAE | MAPE(%) | RMSE | MAE | MAPE(%) | RMSE | MAE | MAPE(%) | RMSE |
| $S_1$ | **P-SSA-TCN-GRU** | **0.0940** | **1.7362** | **0.1238** | **0.2568** | **4.8267** | **0.3265** | **0.4167** | **8.0757** | **0.5294** |
| | P-SSA-GRU | 0.1123 | 3.6966 | 0.2220 | 0.2846 | 5.3506 | 0.3387 | 0.4843 | 9.9426 | 0.6016 |
| | P-SSA-LSTM | 0.0976 | 1.8229 | 0.1240 | 0.2631 | 5.0069 | 0.3321 | 0.5297 | 10.3437 | 0.6508 |
| | P-SSA-RNN | 0.1077 | 2.0292 | 0.1354 | 0.3224 | 6.4276 | 0.3972 | 0.4951 | 10.1777 | 0.6068 |
| | P-SSA-BPNN | 0.2260 | 4.4547 | 0.2761 | 0.3951 | 8.0815 | 0.4885 | 0.5286 | 10.8967 | 0.6623 |
| $S_2$ | **P-SSA-TCN-GRU** | **0.0752** | **1.5659** | **0.1067** | **0.2742** | **5.3426** | **0.3468** | **0.4550** | **8.9857** | **0.5791** |
| | P-SSA-GRU | 0.0853 | 1.9984 | 0.1247 | 0.2934 | 5.7411 | 0.3635 | 0.4867 | 9.9724 | 0.6199 |
| | P-SSA-LSTM | 0.0905 | 1.7425 | 0.1168 | 0.3077 | 5.7009 | 0.3806 | 0.4958 | 10.0417 | 0.6270 |
| | P-SSA-RNN | 0.0872 | 1.6779 | 0.1136 | 0.2834 | 5.4509 | 0.3665 | 0.4982 | 9.9724 | 0.6199 |
| | P-SSA-BPNN | 0.1790 | 3.4107 | 0.2261 | 0.4037 | 8.0571 | 0.5044 | 0.5151 | 10.3526 | 0.6617 |
| $S_3$ | **P-SSA-TCN-GRU** | **0.0943** | **1.7905** | **0.1220** | **0.2620** | **5.0628** | **0.3467** | **0.4252** | **8.1222** | **0.5444** |
| | P-SSA-GRU | 0.1326 | 2.5890 | 0.1656 | 0.3379 | 5.7309 | 0.3791 | 0.5124 | 10.4568 | 0.6531 |
| | P-SSA-LSTM | 0.1106 | 2.2285 | 0.1427 | 0.3317 | 6.6564 | 0.4354 | 0.5336 | 10.8364 | 0.6729 |
| | P-SSA-RNN | 0.1389 | 2.7514 | 0.1861 | 0.3547 | 7.1609 | 0.4569 | 0.5482 | 11.3396 | 0.6847 |
| | P-SSA-BPNN | 0.1730 | 3.4097 | 0.2265 | 0.3982 | 7.9619 | 0.5093 | 0.5510 | 11.3912 | 0.7026 |

### 3.6 Experiment III:Comparison with other TCN-based prediction models

In this experiment, three TCN-based prediction models TCN-LSTM, TCN-RNN, and TCN-BPNN were compared with the proposed model to verify the prediction performance of the model. The prediction results of the proposed model and other TCN-based models are shown in **Fig. 7** and **Table 6**.

In the one-step forecast, the proposed model achieved the highest forecasting accuracy for $S_1$, $S_2$, and $S_3$ wind farms, with MAE index values of 0.0940, 0.0752, and 0.0943, respectively. On the site $S_1$, TCN-LSTM has the lowest prediction accuracy and its MAE is 0.1892, an increase of 0.0949 compared to TCN-GRU. On the site $S_2$, TCN-BPNN has the lowest prediction accuracy, and its MAE is 0.1054, an increase of 0.0302 compared with the proposed model. On the site $S_3$, TCN-BPNN has the lowest prediction accuracy, and its MAE is 0.1583, an increase of 0.064 compared to TCN-GRU. In the two-step advance prediction, the MAPE of the TCN-GRU model is 5.0823%, 5.3426%, and 5.0628% on $S_1$, $S_2$, and $S_3$, respectively. On $S_2$, it is 0.0066%, 0.1117%, and 0.765% lower than the other three models, respectively, and the proposed model has the highest prediction accuracy. In the three-step advance prediction, the RMSEs of the proposed model on the three wind farms were 0.5294, 0.5791, and 0.5444, respectively, maintaining the highest prediction accuracy.

**Remark**: Compared with other TCN-based models, the proposed model has the highest prediction accuracy and the best prediction performance.

## 4 Conclusion

As an important clean and renewable energy, wind power plays an important role in coping with the energy crisis and environmental pollution. However, the volatility and intermittency of wind speed restrict the development of wind power. To improve the utilization rate of wind power, an advanced and high-precision wind speed prediction model P-SSA-TCN-GRU based on data noise reduction algorithm and prediction algorithm is proposed. The model is split into two parts: data denoising and prediction. In the data denoising part, the original wind speed sequence is denoised by the adaptive denoising algorithm P-SSA proposed in this paper, and the interference of the noise on the prediction model gets eliminated. In the prediction part, a new prediction model TCN-GRU is proposed. The receptive field of the samples is expanded by TCN, the wind speed features are strengthened, and the time series information is extracted from the extracted wind speed features by GRU to form a strong wind speed prediction model. Experiment I proved that the P-SSA adaptive noise reduction algorithm proposed is superior to other noise reduction algorithms. Experiment II showed that the proposed TCN-GRU model is superior to other traditional prediction models. Experiment III compared the prediction performance of TCN-GRU and other models based on TCN and verified the best prediction performance of the proposed model. In conclusion, the proposed P-SSA-TCN-GRU model is significantly better than other prediction models, effectively improves the wind speed prediction accuracy, and can be used as a wind speed prediction model for actual wind farms. In the future, the proposed model is hopefully planned to be applied in realizing online real time wind speed prediction.

**Table 6** Prediction results of different TCN-based prediction models on three wind farms

| Site | Model | 1-step | | | 2-step | | | 3-step | | |
|---|---|---|---|---|---|---|---|---|---|---|
| | | MAE | MAPE(%) | RMSE | MAE | MAPE(%) | RMSE | MAE | MAPE(%) | RMSE |
| $S_1$ | **P-SSA-TCN-GRU** | **0.0940** | **1.9042** | **0.1246** | **0.2568** | **5.0823** | **0.3265** | **0.4167** | **8.0757** | **0.5294** |
| | P-SSA-TCN-LSTM | 0.1892 | 3.6966 | 0.2220 | 0.2706 | 5.3506 | 0.3387 | 0.4451 | 9.0361 | 0.5646 |
| | P-SSA-TCN-RNN | 0.1199 | 2.4283 | 0.1601 | 0.2670 | 5.4166 | 0.3304 | 0.4235 | 8.5378 | 0.5365 |
| | P-SSA-TCN-BPNN | 0.1008 | 1.9026 | 0.1251 | 0.2860 | 5.7599 | 0.3563 | 0.4752 | 9.3994 | 0.5863 |
| $S_2$ | **P-SSA-TCN-GRU** | **0.0752** | **1.5659** | **0.1067** | **0.2742** | **5.3426** | **0.3468** | **0.4550** | **8.9857** | **0.5791** |
| | P-SSA-TCN-LSTM | 0.0938 | 1.7584 | 0.1202 | 0.3034 | 5.5492 | 0.3769 | 0.4657 | 9.1524 | 0.6081 |
| | P-SSA-TCN-RNN | 0.0851 | 1.6494 | 0.1124 | 0.2770 | 5.4543 | 0.3527 | 0.4698 | 9.2754 | 0.5884 |
| | P-SSA-TCN-BPNN | 0.1054 | 2.1056 | 0.1342 | 0.3076 | 6.1076 | 0.38 | 0.4771 | 9.4202 | 0.6099 |
| $S_3$ | **P-SSA-TCN-GRU** | **0.0943** | **1.7905** | **0.1220** | **0.2620** | **5.0628** | **0.3467** | **0.4252** | **8.1222** | **0.5444** |
| | P-SSA-TCN-LSTM | 0.1013 | 1.9149 | 0.1312 | 0.2861 | 5.7309 | 0.3791 | 0.4611 | 9.2811 | 0.5953 |
| | P-SSA-TCN-RNN | 0.1334 | 2.7461 | 0.1841 | 0.3099 | 6.1607 | 0.3958 | 0.4653 | 9.3369 | 0.5986 |
| | P-SSA-TCN-BPNN | 0.1583 | 3.0655 | 0.1961 | 0.3694 | 7.3907 | 0.4702 | 0.4839 | 9.8752 | 0.6181 |

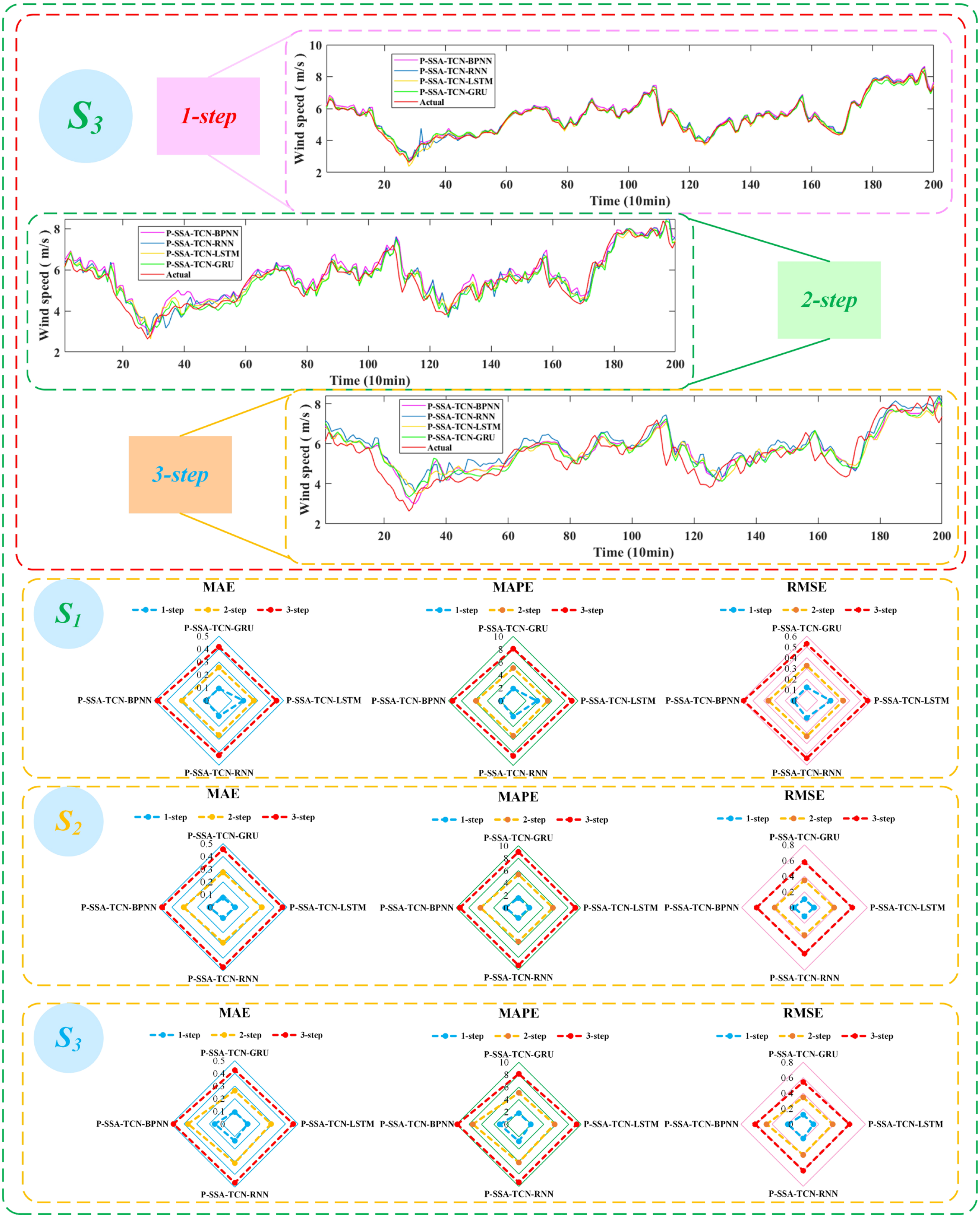

**Fig. 7** The results of different prediction models based on TCN on three wind farms